\documentclass[10pt,twocolumn,letterpaper]{article}

\usepackage[pagenumbers]{cvpr} 

\usepackage{graphicx}
\usepackage{amsmath}
\usepackage{amssymb}
\usepackage{booktabs}
\usepackage{times}
\usepackage{epsfig}
\usepackage{graphicx}
\usepackage{amsmath}
\usepackage{amssymb}
\usepackage{multirow} 
\usepackage{cancel}
\usepackage{rotating}
\usepackage{latexsym}
\usepackage{adjustbox}
\usepackage[thinlines]{easytable}
\usepackage{subcaption}
\usepackage{booktabs}       
\usepackage{mfirstuc}

\usepackage[pagebackref,breaklinks,colorlinks]{hyperref}

\usepackage[capitalize]{cleveref}
\crefname{section}{Sec.}{Secs.}
\Crefname{section}{Section}{Sections}
\Crefname{table}{Table}{Tables}
\crefname{table}{Tab.}{Tabs.}

\def\cvprPaperID{****} 
\def\confName{CVPR}
\def\confYear{2022}

\begin{document}

\title{Robustness and Adaptation to Hidden Factors of Variation}

\author{William Paul\\
Johns Hopkins University Applied Physics Laboratory\\
11100 Johns Hopkins Road Laurel, Maryland 20723\\
{\tt\small william.paul@jhuapl.edu}
\and
Philippe Burlina\\
Johns Hopkins University Applied Physics Laboratory\\
11100 Johns Hopkins Road Laurel, Maryland 20723\\
{\tt\small philippe.burlina@jhuapl.edu}
}
\maketitle

\begin{abstract}
    We tackle here a specific, still not widely addressed aspect, of AI robustness, which consists of seeking invariance / insensitivity of model performance to hidden factors of variations in the data. Towards this end, we employ  a two step strategy that a) does unsupervised discovery, via generative models, of sensitive factors that cause models to under-perform, and b) intervenes models to make their performance invariant to these sensitive factors' influence. We consider 3 separate interventions for robustness, including: data augmentation, semantic consistency, and adversarial alignment. We evaluate our method using metrics that measure trade offs between invariance (insensitivity) and overall performance (utility) and show the benefits of our method for 3 settings (unsupervised, semi-supervised and generalization).

\end{abstract}
\vspace{-0.5cm}
\section{Introduction}

{\bf Motivation} Model robustness and AI assurance  encompass many facets, including resilience to adversarial attacks~\cite{carlini2017adversarial}, domain adaptation~\cite{visda2021}, and have extended recently to broader problems such as fairness~\cite{paul2020tara}, or privacy~\cite{shokri2017membership}. This paper explores an empirical approach to probing robustness, which is understood here as invariance/insensitivity of the model performance to innate and possibly unknown variations in the data. This work  relates to domain adaptation and generalization, as those may arguably be enhanced by desensitizing the model to hidden factors of variation. In a sense, this works also generalizes robustness as addressed in fairness studies, which - in contrast to our setting -- probes invariance and equality of performance with regard to protected factors that are known a-priori (e.g. gender in face images). 

Indeed, our motivation here is real-life scenarios where incomplete a-priori knowledge exists about important factors of variation that may cause models to fail (e.g. for pedestrian or traffic sign detection, or face recognition). This leads us to consider various settings (unsupervised,  semi-supervised, generalization) that correspond to different degrees of prior knowledge or available labels about these factors. 

Indeed, two challenges exist: one is related to the cost of collecting/curating more more annotated data to evaluate and mitigate this type of invariance to factors of variation; and the other lies in the possible lack of knowledge regarding which factors of variation actually impact performance, and the fact that often, not all factors are known, or, if some factors are known, labels may be unavailable or scarce. These limitations motivate the need to consider various unsupervised or generalization settings herein. 

{\bf Approach:} Our response to those issues is to develop a general purpose, dual-pronged approach that: 1) leverages generative models that can both generate synthetic data and discover in an unsupervised fashion sensitive factors of variation, and 2)  intervenes models vis-a-vis those factors. 
Specifically our approach is as follows (see Figure \ref{fig:pipeline}): we first learn factors $\{C_i\}$ that correlate (in the mutual information sense) to variations in the data. Those $C_i$ are directions in latent space of InfoStyleGAN\cite{paul2020unsupervised}. We then evaluate sensitivity/invariance, i.e., how influential each $C_i$ is on task performance. We then enhance robustness by intervening directly on the task model sensitivity for each $C_i$. We compare individually three interventions: data augmentation, semantic consistency, and adversarial alignment.

\section{Prior Work} 
Taking a broad, conceptual view of robustness, our approach is most related to approaches in robustness concerned about domain adaptation, (in particular methods using adversarial alignment and style transfer), and robustness methods concerned with fairness.  

{\bf Generative Methods:} those methods, when used for robustness taken in the sense of adaptation, alter training data to facilitate pseudo-labeling, or for source-to-target domain translation. They use techniques such as variational autoencoders (VAEs)~\cite{madras2018learning,kingma2013auto, louizos2015variational}; GANs~\cite{karras2019style,grover2019bias,zhao2017infovae}; or CycleGAN~\cite{cyclegan2017}, StarGAN~\cite{stargan2018} or pix2pix~\cite{pix2pix2017, pix2pix2019}. Generative methods when applied to robustness in the sense of fairness,  include~\cite{quadrianto2019discovering,hwang2020fairfacegan,sattigeri2018fairness}, and are used for augmentation to address data imbalance (a situation akin to prior shift for domain generalization). Our aims here are to seek generative methods that are able to discover (for analysis) and control (for synthesis) factors of variation in data, where the aforementioned techniques are not appropriate, in that they fail to allow fine control and disentanglement of factors of variations; 
Or are only applicable when factors are  known a-priori. Instead, we leverage a type of generative model recently developed (InfoStyleGAN~\cite{paul2020unsupervised}) that address both such needs for analysis and synthesis, by discovering, de-novo, and then controling for, unknown factors of variation within  data that may affect robustness, understood here as invariance / insensitivity of the performance of models vis-a-vis the variation in such factors.

{\bf Adversarial Approaches} these are used to intervene on  models to achieve source / target domain feature alignment in domain adaptation~\cite{ganin2016domain}. Related methods have been applied to fairness where alignment is done vis as vis protected factors~\cite{beutel2017data,wadsworth2018achieving,zhang2018mitigating,alvi2018turning,song2019learning}.

Compared to prior work, our novel contributions are:
\begin{enumerate}
    \item {\em Achieving Model Invariance With Regard to Unknown Sensitive Factors of Variation} We expand the scope of domain adaptation and fairness,  to more general settings of arbitrary factors, and importantly, we consider these factors are not known a-priori and  use generative models to discover them.
    \item {\em Unsupervised / Generalization / Semi-Supervised Settings} We consider different settings corresponding to different degrees of knowledge available about these sensitive factors. We address novel settings with no knowledge (unsupervised), some knowledge (semi-supervised) and ask the question how discovered factors may generalize from source to target factors, which is a specific type of domain generalization. 
    \item {\em Interventions} We evaluate the efficacy of achieving robustness and invariance vis-a-vis those factors, by comparing three types of interventions including augmentation,  semantic consistency and adversarial interventions. And, in semi-supervised settings, we propose a hyperparameter selection method to select which combination of factors to intervene, and which interventions to apply, to achieve optimal outcomes, via leveraging validation data.
\end{enumerate}

\section{Methods}
\begin{figure*}
\centering 
\includegraphics[page=2,width=14cm]{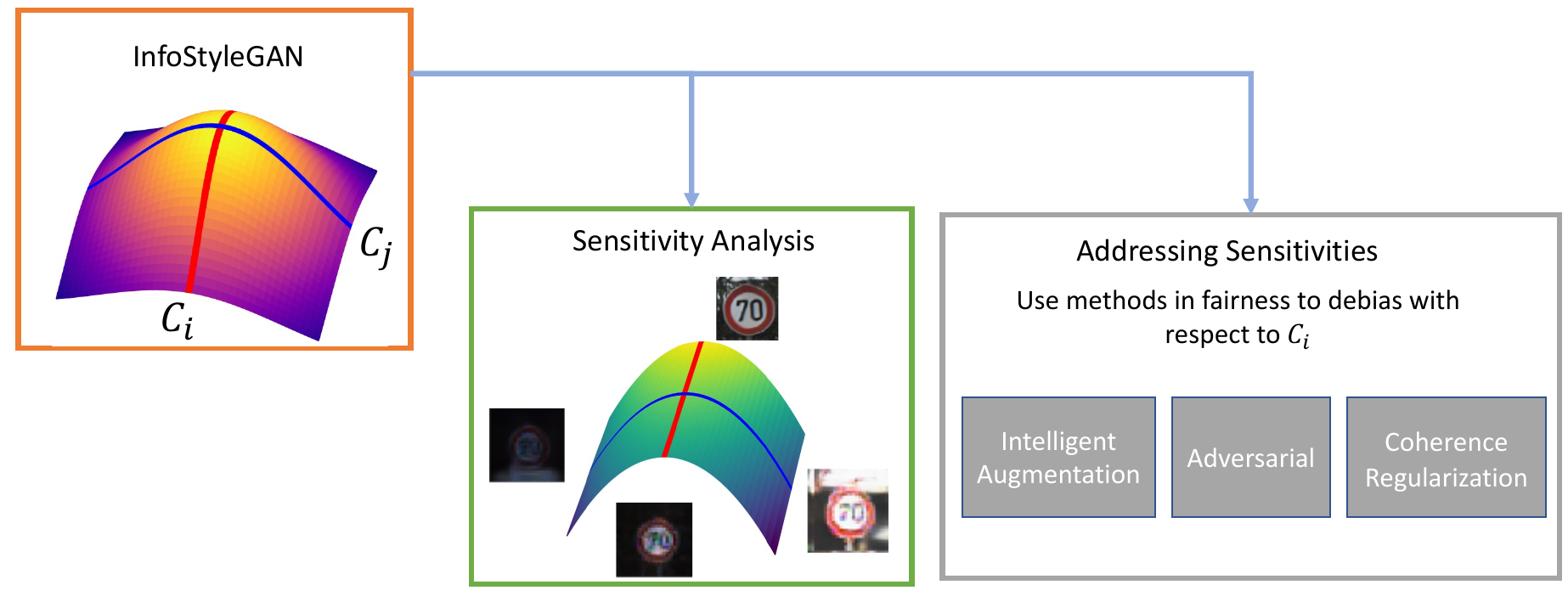}
     \vspace{-0.01in}
    \caption{Overall Approach: 1) We use a generative model to discover and control factors of variation in the data. This uses mutual information to find latent space variables aligned with factors of variations, similar to PCA. 2) For those factors found to influence most the end task performance, we use three types of interventions on the end-task training data and model, to promote performance invariance / robustness, vis a vis those most influential factors, including: data augmentation (sampling more data along the latent directions of variation); semantic consistency; and adversarial alignment. }
\label{fig:pipeline}
\end{figure*}

Our overall pipeline is  a two-pronged approach, first to discover factors of variation, then to perform interventions, and is summarized in Figure \ref{fig:pipeline}. 

\textbf{Discovery of Factors of Variation:}  
In this work, we proceed by positing the existence of yet unknown  factors $\{A_i\}$ present in images ${X}$ that could influence the behavior of a model/predictor. Consequently, our goal is to recover $\{A_i\}$ via a generative model's latent variables $\{C_i\}$, and then evaluate the invariance of each $C_i$. One caveat is that relating some unknown $A_i$ with a $C_i$ is difficult if not impossible without strong inductive biases \cite{Locatello2019OnTF}.
To address this problem, and in order to recover $\{A_i\}$ in variables $\{C_i\}$, we follow the approach in \cite{paul2020unsupervised}. This technique builds on  \cite{zhao2017infovae} by augmenting the latent codes $\{C_i\}$ used in conjunction with a StyleGAN architecture. Consequently, each individual $C_i$ can now influence the image generation at multiple scales, via maximizing mutual information between latent factors and generated images, thereby also allowing for better control over potential attributes in the synthetic image. In practice, in order to ensure $C_i$ aligns with potential factors of variation $A_i$ in images, we require that $C_i$ be reconstructed from the image, via an auxiliary network $Q$ that predicts $C_i$ from the generated image~\cite{paul2020unsupervised}. We take $Q$ to share weights with the discriminator $D$, where $Q$ takes as input an intermediate representation from $D$. To dissociate  $C_i$ and $Y$ influence on the generation of images, we also make the generator conditional on $Y$. This loss function is used:
\begin{align}
L_{\text{in}} = \mathbb{E}_{\hat{X} \sim G(Z, C, Y)}(\mathbb{CE}(C, Q(\hat{X})) - \log p(C))
\end{align}

\noindent is then added to the loss function for $G$ and $Q$, where $\mathbb{E}(\cdot)$ denotes expectation and $\mathbb{CE}(\cdot;\cdot)$ the cross entropy.

Consequently, the full scheme for the GAN is:
\begin{align}
    \min_{G, Q} \max_D \mathbb{E}_{X}{\log D(X)} + \mathbb{E}_{\hat{X}}(\log (1 - D(\hat{X}))+ L_{\text{in}}
\end{align}
{\bf Characterizing Invariance and Mitigating Sensitivities:}
Once $C_i$ have been discovered, we can then easily evaluate invariance, by estimating how much changes in $C_i$ impact changes in task performance. And mitigate any sensitivity to variations in these factors using  interventions on the training data or the task prediction model, including augmenting the training data to better reflect variations entailed by changing these factors; or using semantic consistency to variations in these factors; or  adversarial alignment to make prediction independent of these factors.

\textbf{Interventions to Achieve Invariance to Factors of Variation:} We next present three methods for mitigating sensitivities/achieving invariance. These interventions are used and evaluated independently in this work, and their joint combination is left for future work.

\textit{Data Augmentation (DA):} The idea here is to augment the training data with more data sampled in latent space along the direction corresponding to the factor of variation we want to become invariant against.
We sample an additional fixed size dataset from the generator, and use these data as supervised examples augmenting the original dataset. To achieve invariance for a specific factor $C_i$, we sample $C_i$ from a uniform distribution from -2 to 2 to ensure equal coverage across the learnt hidden factor with the aim of not impacting image realism, akin to the truncation trick in~\cite{brock2019large}. An alternative not used herein is to sample from heavy tailed distributions to probe long tail/edge case exemplars corresponding to these factors of variation.

\textit{Adversarial Alignment (AA):} We intervene directly the task model here, to make it ``blind'' to the knowledge of the sensitive factor label, hoping therefore to make the task prediction invariant to the sensitive factor, by employing an adversary to predict the sensitive factor from an embedded representation of the image we want to render ``blind'', as is done in other works such as \cite{zhang2018mitigating,ganin2016domain}. The adversary is given the target label (as each label may have different levels of invariance to the sensitive factor) and the embedding taken after the global average pooling layer. The adversary and classifier are trained in a alternating optimization scheme.

\textit{Semantic Consistency (SC):} As an alternative to the model intervention that actively defends against an adversary predicting the sensitive factor, we can instead take as inductive bias the view that the classifier should remain semantically consistent no matter the variation in sensitive factors, i.e., output the same probabilities regardless of these factors' values. For the sake of argument and taking as an example where  the learnt factor were to correspond to a known factor, say ``presence of glasses'', for a face verification task, then changing this factor from true to false ought not change the predictions of the classifier doing face recognition. Taking inspiration from \cite{xie2020unsupervised} and \cite{dash2020evaluating}, we enforce this semantic consistency constraint by adding a KL divergence between the same image with two different values for the protected factor. Note that $\hat{X}_{C_i \leftarrow c_i'}$ denotes taking the (Z,C, Y) that produced $\hat{X}$, setting $C_i$ to $c_i'$ and generating the corresponding image.
Thus, the full loss term for the classifier $F$ becomes:
\begin{align}
    L_{SC}(F) &= \mathbb{E}_{Z,C,Y}\mathbb{KL}(\hat{Y}(\hat{X}_{C_i \leftarrow c_i}) || \hat{Y}(\hat{X}_{C_i \leftarrow c_i'})))
\end{align}

\section{Settings} 
We consider three settings for evaluating outcomes:
\begin{enumerate}
\item {\bf Unsupervised Factor Invariance}: this is the native setting of this work where we only assume we have  images and labels for the end task, but no knowledge on the sensitive factors, forcing us to discover sensitive factors and intervene on these. 
\item {\bf Factor Generalization:} this setting tests generalization via evaluation on real factors, probing the question: do interventions derived via unsupervised setting 1. achieve a more robust end-task model, and yield invariance, when applied and tested on real factors they did not have access to originally? 
\item {\bf Semi-supervised}: In an extension of setting 2., we assume here we have available some data examples with labeled known factors, which we then use as validation data for selection of interventions.  We perform  automated selection of hyperparameters of interventions found in setting 1., consisting of choosing which combination of sensitive factor, and intervention method, is best apt at desensitizing/robustifying  some known validation factors. This setting a type of ({\em semi-supervised}) adaptation from source/discovered unknown factors to target/known factors.  We refer to this last approach as ``ACAI'' (for ``data Augmentation, semantic Consistency, and Adversarial alignment Intervention'').
\end{enumerate}
\section{Experiments}
\textbf{Datasets:} We evaluate on two problem domains / image datasets: One is traffic sign classification using the German Traffic Sign Recognition Benchmark (GTSRB) \cite{Stallkamp-IJCNN-2011}; and face analytics using CelebA \cite{liu2015faceattributes}.

German Traffic Sign Recognition Benchmark is an image dataset containing 43 different classes of traffic signs. Common factors  present in this dataset include the size of the sign and the lighting condition, which can affect how finer details appear in the image. The image size is 64 pixels by 64 pixels, and we use 45,322 images for training and validation, and 10,112 images for testing. The synthetic dataset, correspondingly, is 101,120 images. The ground truth factors we use for this when testing the generalization and semi-supervised settings are the sign area, which is the area computed from the sign ROI normalized to be between 0 and 1, and the computed brightness, considered as the luminance in CIELab space averaged over the image.

CelebA is a face image dataset consisting of celebrities, with many different attributes labeled ranging from the gender to the attractiveness of the person. We predict the age of the person. The images are cropped to be 128 pixels by 128 pixels, and we use 162,121 images for training, and 39,201 images for testing. The ground truth factors we use for this dataset for the generalization and semi-supervised setting, are  skin color, which is the ITA computed over the face as in \cite{paul2020tara}, and the computed brightness of image, computed as in GTSRB above.

The factor variation``subpopulations'' (i.e. subsets) for all ground truth factors (except ITA for CelebA) are taken by binning (10 bins), where $A_i(j)$ or $C_i(j)$ denotes the jth bin for the ith true or discovered factor respectively. For ITA, there are only two subpopulations that are split using a threshold of 17, which was found in \cite{paul2020tara} to almost exclude facial images that denoted as 'Pale Skin'.

\textbf{Experiments:} For each dataset, we first partition into a training dataset and a test dataset. The training dataset is used to both train InfoStyleGAN as well as the task classifiers. The test dataset is used to evaluate the overall invariance and utility metrics for every method, and is also used to compute invariance metrics in the generalization and semi-supervised settings where test labels for ground truth factors are available.

We first train InfoStyleGAN on the dataset by dedicating ten $C_i$ variables, to discover sensitive factors innate in the data. We train a baseline task classifier, which is a ResNet50 pre-trained on ImageNet, on the given task for the dataset.  

For the {\bf unsupervised setting}, we  intervene on all sensitive factors for the baseline and evaluate the resulting change in utility and invariance metrics (defined in a later section). We compare and evaluate four models (the baseline model and the baseline intervened with the three mitigations we described in the methods section). This evaluation is done on synthetic images as follows: we sample data conditioned on the class labels from InfoStyleGAN where each $C_i$ is sampled uniformly from -2 to 2 for the same reasons as the augmentation method. We partition the resulting images according to the image's $C_i$, where there are 10 bins between -2 and 2, leading to 10 variations/subpopulations for the same factor. This synthetic dataset has the same task label distribution as the test set, and is sampled to be ten times the size, so each partition is roughly equal size to the original test set. We then compute the overall accuracy for each image partition, and treat these individual partitions as subpopulations to compute the invariance metrics.

For the {\bf generalization setting} the high level goal is to assess how well an intervention developed under the unsupervised setting actually performs when generalized and applied to a specific / real sensitive factor for which we now have ground truth. Therefore testing is here done for a factor that appears to align semantically with a discovered factor of variation. For this setting we report the resulting performance corresponding to the baseline model and to all three types of interventions. 

For the {\bf semi-supervised setting}, we instead assume we now have access to a validation dataset of data with labels for the real sensitive factor used in the previous setting, and that this data is now  used to perform hyperparameter selection dictating the combination of which intervention to apply and which factor to intervene for optimal outcomes. When we have the validation labels, we essentially do a grid search for those hyperparameters; we intervene on every $C_i$ and use the validation labels to select the best pair of $C_i$ and intervention type. For nomenclature sake, we denote the selected intervention as Intervention-i, e.g.  DA-1 denotes the data augmentation on $C_1$. All results for all settings are reported in the same tables for a given sensitive factor.

\begin{table}[t]
\includegraphics[width=\linewidth]{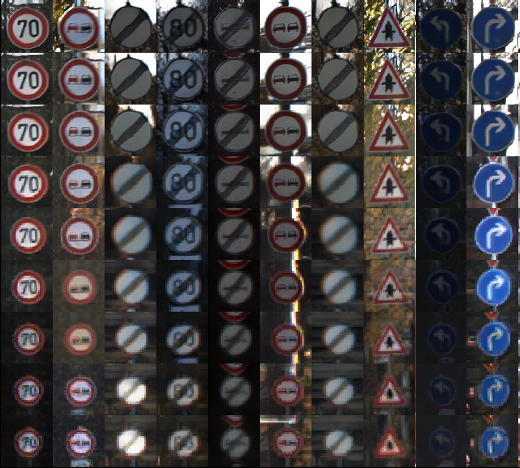}

\centering
\scriptsize
    \centering
    	\begin{tabular}{c|c|c|c|c|c|c}
	    \toprule
        Setting&Interv.&Acc&$\text{Acc}_{\text{gap}}$&$\text{Acc}_{\text{min}}$& $\text{CAI}_{0.5} $  &  $\text{CAI}_{0.75}$  \cr
        \midrule
        \midrule
        \multirow{4}{*}{\parbox{1.0cm}{\centering \textbf{Unsup. Invariance}}}&Base &98.49&22.93&65.80&0.00&0.00 \cr
        &DA-1 &\textbf{98.66}&\textbf{4.12}&79.31&\textbf{9.49}&\textbf{14.17} \cr 
        &AA-1 &97.99&10.84&\textbf{86.09}&5.80&8.96 \cr
        
        &SC-1 &96.31&34.85&48.50&-7.05&-9.52 \cr \cline{1-7}
        \midrule
        \multirow{4}{*}{\parbox{1.0cm}{\centering \textbf{Factor Generalization}}}&Base &98.49&3.33&96.24&0.00&0.00 \cr
        &DA-1 &\textbf{98.66}&\textbf{2.83}&\textbf{96.82}&\textbf{0.34}&\textbf{0.42} \cr 
        &AA-1 &97.99&\textbf{2.83}&96.49&0.00&0.25 \cr
        
        &SC-1 &96.31&5.56&92.76&-2.20&-2.22 \cr \cline{1-7}
        \multirow{3}{*}{\parbox{1.0cm}{\centering \textbf{Semisup. Invariance}}}&\multirow{3}{*}{\parbox{0.8cm}{ACAI (DA-4)}} &\multirow{3}{*}{\textbf{99.12}}&\multirow{3}{*}{\textbf{1.42}}&\multirow{3}{*}{\textbf{98.15}}&\multirow{3}{*}{\textbf{1.27}}&\multirow{3}{*}{\textbf{1.59}} \cr
        &&&&&\cr
        &&&&&\cr
        \bottomrule
	\end{tabular}
	
    \captionof{table}{GTSRB Image transitions and results for the most influential discovered factor (found as sensitive Factor 1, out of 10 sensitive factors found) which visually appears to correspond to ``Sign Size''. Invariance metrics for the unsupervised setting are computed with respect to this, and invariance metrics for the generalization and semisupervised setting are therefor computed with respect to the ``true sign area'' factor.}
    \label{tab:gtsrb1}
\end{table}
\begin{table}[t]

\includegraphics[width=\linewidth]{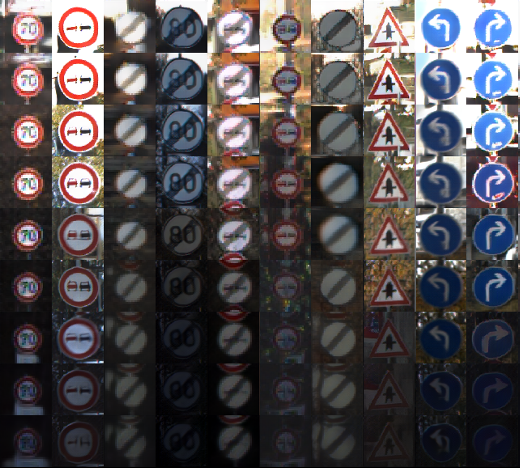}

\scriptsize
    \centering
    	\begin{tabular}{c|c|c|c|c|c|c}
	    \toprule
        Setting&Interv.&Acc&$\text{Acc}_{\text{gap}}$&$\text{Acc}_{\text{min}}$& $\text{CAI}_{0.5}$  &  $\text{CAI}_{0.75}$  \cr
        \midrule
        \midrule
        \multirow{4}{*}{\parbox{1.0cm}{\centering \textbf{Unsup. Invariance}}}&Base &98.49&18.46&69.37&0.00&0.00 \cr
        &DA-7 &\textbf{99.06}&\textbf{2.81}&\textbf{95.82}&\textbf{8.09}&\textbf{11.87} \cr 
        &AA-7 &98.01&4.45&92.00&6.74&10.38 \cr
        
        &SC-7 &95.96&15.65&65.30&0.10&1.46 \cr \cline{1-7}
        \midrule
        \multirow{4}{*}{\parbox{1.0cm}{\centering \textbf{Factor Generalization}}}&Base &98.49&\textbf{2.10}&\textbf{97.37}&\textbf{0.00}&\textbf{0.00} \cr
        &DA-7 &\textbf{99.06}&3.32&96.50&-0.33&-0.78 \cr 
        &AA-7 &98.01&3.13&95.98&-0.75&-0.89 \cr
        
        &SC-7 &95.96&3.81&94.09&-2.12&-1.91 \cr \cline{1-7}
        \multirow{3}{*}{\parbox{1.0cm}{\centering \textbf{Semisup. Invariance}}}&\multirow{3}{*}{\parbox{0.8cm}{ACAI (DA-9)}} &\multirow{3}{*}{\textbf{98.93}}&\multirow{3}{*}{\textbf{1.67}}&\multirow{3}{*}{\textbf{97.81}}&\multirow{3}{*}{\textbf{0.44}}&\multirow{3}{*}{\textbf{0.43}} \cr
        &&&&&\cr
        &&&&&\cr
        \bottomrule
	\end{tabular}
	
    \captionof{table}{GTSRB Image transitions and results for discovered sensitive Factor 7 (out of 10) which visually appears to correspond to actual factor ``lighting''.}
    
    \label{tab:gtsrb7}
\end{table}
\textbf{Utility/Invariance Metrics} As  interventions on the data and model to achieve sensitive factor invariance may inevitably lead to reduced overall accuracy/utility for the end classification task, we report metrics that characterize the two end goals (utility and invariance).   Overall accuracy is used to characterize model utility. For evaluating invariance, we compute the accuracy gap over the variations in the factor $A$ or $C_i$ to be the maximum overall accuracy over the variations in this factor, minus the minimum overall accuracy:
\begin{align}
    \text{Acc}_{\text{gap}}(\hat{Y}) = \max_{j} \text{Acc}(\hat{Y}(\hat{X}_{C_i(j)})) \nonumber \\ 
    - \min_{j} \text{Acc}(\hat{Y}(\hat{X}_{C_i(j)}))
\end{align}
for a classifier $\hat{Y}$ where $\hat{X}_{C_i(j)}$ denotes the synthetic imagery corresponding with the jth bin of $C_i$. $X_{A_i(j)}$ is used for evaluation on real imagery. We also compute the minimum accuracy over the subpopulations, which aligns well with observing the worst case performance.

As an important question is characterizing jointly the trade-off in invariance and utility, we also compute a {\em compound accuracy improvement} metric we call $\text{CAI}_\lambda (\hat{Y})$ as in \cite{paul2020tara}, which is the weighted sum of the improvement in utility (as measured via the difference in accuracy between the intervened model for classifying $\hat{Y}$ and the baseline model), and the improvement in invariance (as measured via the decrease in accuracy gap from the intervened and the baseline model for classifying $\hat{Y}$).
\begin{align}
    \text{CAI}_{\lambda}(\hat{Y}) = &\lambda (\text{Acc}_{\text{gap}}(\text{baseline}) - \text{Acc}_{\text{gap}}(\hat{Y})) + \nonumber \\
    &(1-\lambda ) (\text{Acc}(\hat{Y}) - \text{Acc}({\text{baseline}}))
\end{align}
\textbf{Results:} We show the results of intervening on the two most sensitive factors for the baseline in the unsupervised setting for both the CelebA and the traffic sign experiment herein.

We show those results in tables \ref{tab:gtsrb1}, \ref{tab:gtsrb7}, \ref{tab:celeba5}, and \ref{tab:celeba4}. Each table has a companion figure above it which corresponds to alterations of the image resulting from varying the factor value in latent space, with  variations shown along the rows of the figure, and where each column of the figure corresponds to an different image taken.  For the table itself, each numerical column corresponds to a specific metric used, where the invariance metrics for unsupervised invariance are computed over $C_i$ and the invariance metrics for the other two settings are over actual factor $A$ the discovered factor appears to correspond to. 

\section{Discussion}
\begin{table}[]
\includegraphics[width=\linewidth]{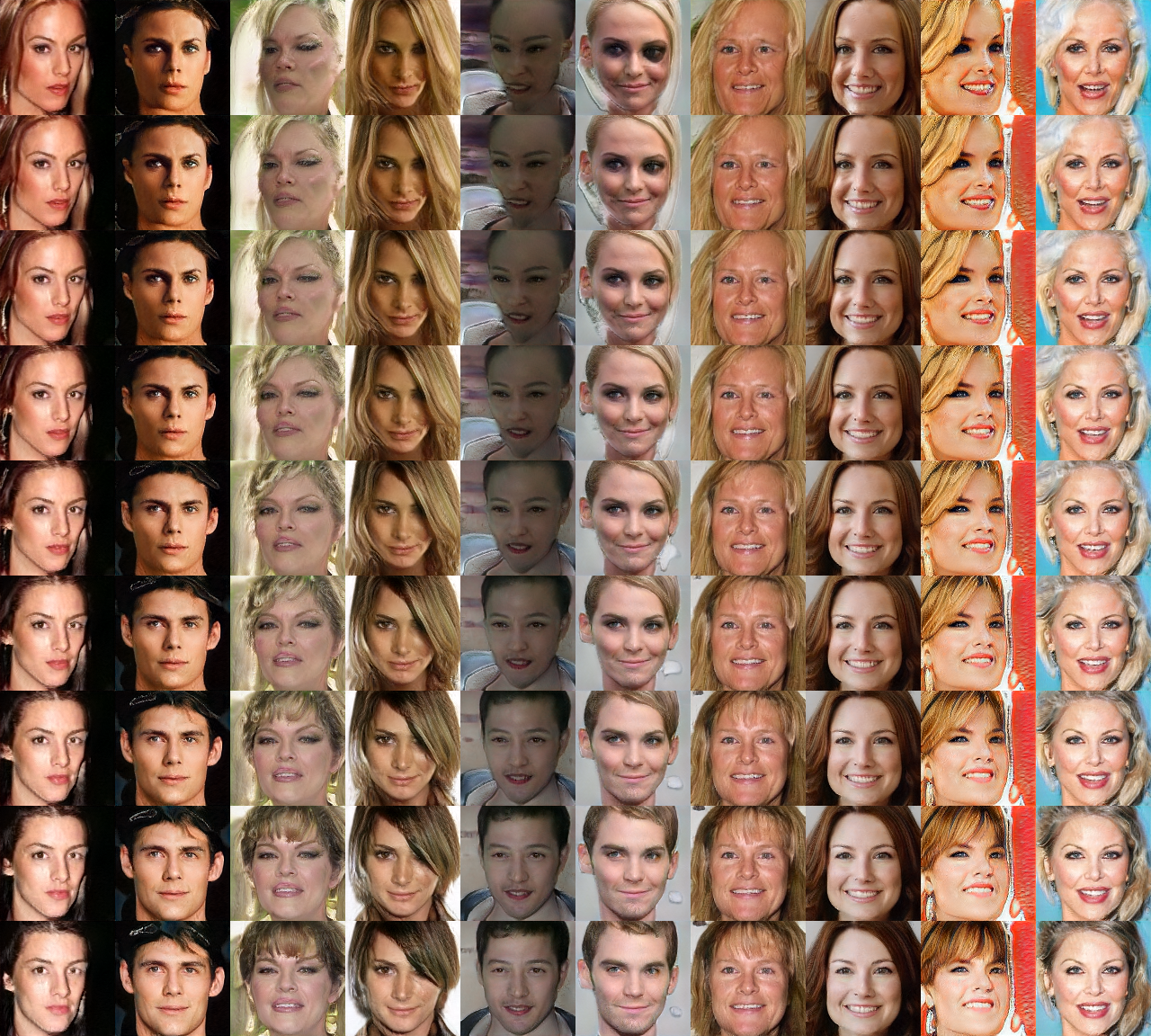}

\scriptsize
    \centering
    	\begin{tabular}{c|c|c|c|c|c|c}
	    \toprule
        Setting&Interv.&Acc&$\text{Acc}_{\text{gap}}$&$\text{Acc}_{\text{min}}$& $\text{CAI}_{0.5}$  &  $\text{CAI}_{0.75}$  \cr
        \midrule
        \midrule
        \multirow{4}{*}{\parbox{1.0cm}{\centering \textbf{Unsup. Invariance}}}&Base &\textbf{81.34}&8.77&69.80&0.00&0.00 \cr
        &DA-5 &79.45&\textbf{1.02}&\textbf{86.34}&\textbf{2.93}&\textbf{5.30} \cr 
        &AA-5 &74.11&1.36&84.37&0.15&3.79 \cr
        
        &SC-5 &80.38&7.86&67.94&0.02&0.47 \cr \cline{1-7}
        \midrule
        \multirow{4}{*}{\parbox{1.0cm}{\centering \textbf{Factor Generalization}}}&Base &\textbf{81.34}&6.07&78.20&0.00&0.00 \cr
        &DA-5 &79.45&6.86&\textbf{79.93}&-1.34&-1.06 \cr 
        &AA-5 &74.11&10.41&67.96&-5.78&-5.06 \cr
        
        &SC-5 &80.38&\textbf{4.64}&78.08&\textbf{0.23}&\textbf{0.83} \cr \cline{1-7}
        \multirow{3}{*}{\parbox{1.0cm}{\centering \textbf{Semisup. Invariance}}}&\multirow{3}{*}{\parbox{0.8cm}{ACAI (DA-2)}} &\multirow{3}{*}{\textbf{82.53}}&\multirow{3}{*}{\textbf{5.41}}&\multirow{3}{*}{\textbf{79.46}}&\multirow{3}{*}{\textbf{0.92}}&\multirow{3}{*}{\textbf{0.79}} \cr
        &&&&&\cr
        &&&&&\cr
        \bottomrule
	\end{tabular}
	
    \captionof{table}{CelebA Image transitions and results for sensitive factor 5 (most sensitive) which visually appears to correspond to lighting. Invariance metrics for the unsupervised setting is computed with respect to factor 5, and invariance metrics for the other two settings are computed with respect to the true lighting factor.}
    \label{tab:celeba5}
\end{table}
\begin{table}[]

\includegraphics[width=\linewidth]{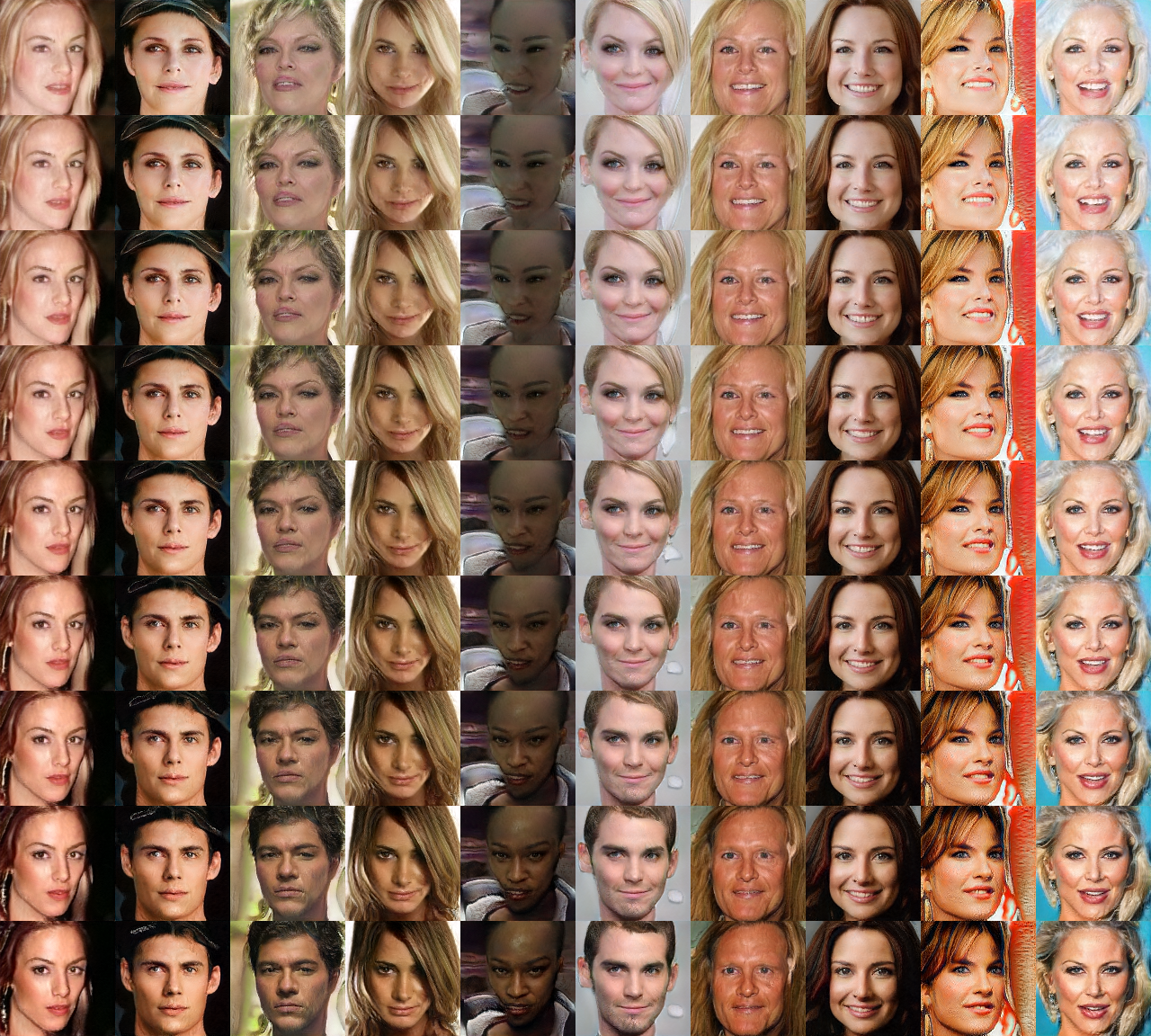}

\scriptsize
    \centering
    	\begin{tabular}{c|c|c|c|c|c|c}
	    \toprule
        Setting&Interv.&Acc&$\text{Acc}_{\text{gap}}$&$\text{Acc}_{\text{min}}$& $\text{CAI}_{0.5}$  &  $\text{CAI}_{0.75}$  \cr
        \midrule
        \midrule
        \multirow{4}{*}{\parbox{1.0cm}{\centering \textbf{Unsup. Invariance}}}&Base &81.34&7.20&70.09&0.00&0.00 \cr
        &DA-4 &\textbf{82.53}&\textbf{1.05}&\textbf{86.19}&\textbf{3.72}&\textbf{4.92} \cr 
        &AA-4 &80.12&3.99&81.59&0.99&2.10 \cr
        
        &SC-4 &81.48&2.12&69.48&2.58&3.83 \cr \cline{1-7}
        \midrule
        \multirow{4}{*}{\parbox{1.0cm}{\centering \textbf{Factor Generalization}}}&Base &81.34&5.27&76.83&0.00&0.00 \cr
        &DA-4 &\textbf{82.53}&5.32&77.96&0.57&0.26 \cr 
        &AA-4 &80.12&\textbf{2.86}&77.67&0.59&\textbf{1.50} \cr
        
        &SC-4 &81.48&3.96&\textbf{78.09}&\textbf{0.72}&1.02 \cr \cline{1-7}
        \multirow{3}{*}{\parbox{1.0cm}{\centering \textbf{Semisup. Invariance}}}&\multirow{3}{*}{\parbox{0.8cm}{ACAI (SC-1)}} &\multirow{3}{*}{\textbf{81.94}}&\multirow{3}{*}{\textbf{3.07}}&\multirow{3}{*}{\textbf{79.31}}&\multirow{3}{*}{\textbf{1.40}}&\multirow{3}{*}{\textbf{1.80}} \cr
        &&&&&\cr
        &&&&&\cr
        \bottomrule
	\end{tabular}
    \captionof{table}{CelebA Image transitions and results Sensitive Factor 4 (second most influential out of 10 discovered factors) which alsp visually appears to correspond to skin color.}
    \label{tab:celeba4}
\end{table}
\textbf{Unsupervised Invariance:} For this setting, we see that the initial high sensitivity to the factor (high accuracy gaps) for the model with no intervention are mitigated significantly by intervening on that sensitive factor. Typically, intervening also slightly improved the overall accuracy on real factors, except for Table \ref{tab:celeba5} for lighting, where the highest accuracy over the interventions is 80.38\% for CR-5. Due to this, the methods that the CAI metrics were highest at, were also the ones that had the lowest accuracy gap.

Data Augmentation was the best performing intervention for all metrics except for Table \ref{tab:celeba4} where the SC method has a better overall accuracy and $\text{CAI}_\text{0.5}$. For the GTSRB dataset, the AA method came in second compared to DA and still improved over the baseline. SC is worse than the other two interventions, and worse than the baseline when addressing sign size. For CelebA, AA exhibited worse overall accuracies compared to SC. SC had the second best accuracy gap for skin color, and the second best accuracy and second worst accuracy gap for lighting.

\textbf{Factor Generalization:} For GTSRB, we see that the DA and AA interventions improve upon the base classifier for sign area in terms of both the accuracy gap as well as the minimum accuracy. However, no intervention is best in terms of invariance for lighting in this setting. Of the interventions, data augmentation performed best in terms of $\text{CAI}_{0.5}$.
For CelebA, SC had the best $\text{CAI}_{0.5}$, having the best accuracy gap for lighting and best overall accuracy for skin color. Only SC improved upon the original model for lighting, but all three interventions improved in either overall accuracy or the accuracy gap for skin color.

\textbf{Semi-supervised Invariance:} Due to intervening over any factor and approach, the intervention methods have the highest $\text{CAI}_{0.5}$ over all interventions evaluated on real factors. For GTSRB, ACAI primarily improved the accuracy gap, whereas for CelebA ACAI has a slight decrease in the accuracy gap and an increase in the overall accuracy.

\textbf{Limitations:} 
The model we use, and most approaches for discovering factors, focus on more 'global' factors that affect the image completely, as we see in the factors our InfoStyleGAN models capture. More complex scenes such as those found in autonomous driving or aerial imagery have 'local' factors that affect only single entities in scenes, such as vehicles or people. To the best of our knowledge, capturing local factors has not been researched extensively.

\section{Conclusions}

We demonstrate the benefits of a new approach for robustness that uses a two prong {\em }discover and intervene strategy, for factors of variations in the data that seem to most influence variations in performance for the end-task.

{\small
\bibliographystyle{ieee_fullname}
\bibliography{egbib}

\begin{thebibliography}{10}\itemsep=-1pt

\bibitem{alvi2018turning}
Mohsan Alvi, Andrew Zisserman, and Christoffer Nell{\aa}ker.
\newblock Turning a blind eye: Explicit removal of biases and variation from
  deep neural network embeddings.
\newblock In {\em Proceedings of the {E}uropean {C}onference on {C}omputer
  {V}ision ({ECCV}) {W}orkshops}, 2018.

\bibitem{visda2021}
Dina Bashkirova, Dan Hendrycks, Donghyun Kim, Samarth Mishra, Kate Saenko,
  Kuniaki Saito, Piotr Teterwak, and Ben Usman.
\newblock Visda-2021 competition universal domain adaptation to improve
  performance on out-of-distribution data, 2021.

\bibitem{beutel2017data}
Alex Beutel, Jilin Chen, Zhe Zhao, and Ed~H Chi.
\newblock Data decisions and theoretical implications when adversarially
  learning fair representations.
\newblock {\em arXiv preprint arXiv:1707.00075}, 2017.

\bibitem{brock2019large}
Andrew Brock, Jeff Donahue, and Karen Simonyan.
\newblock Large scale gan training for high fidelity natural image synthesis,
  2019.

\bibitem{carlini2017adversarial}
Nicholas Carlini and David Wagner.
\newblock Adversarial examples are not easily detected: Bypassing ten detection
  methods.
\newblock In {\em Proceedings of the 10th {ACM} {W}orkshop on {A}rtificial
  {I}ntelligence and {S}ecurity}, pages 3--14, 2017.

\bibitem{stargan2018}
Yunjey Choi, Minje Choi, Munyoung Kim, Jung-Woo Ha, Sunghun Kim, and Jaegul
  Choo.
\newblock Stargan: Unified generative adversarial networks for multi-domain
  image-to-image translation.
\newblock In {\em Proceedings of the {IEEE} {C}onference on {C}omputer {V}ision
  and {P}attern {R}ecognition}, June 2018.

\bibitem{dash2020evaluating}
Saloni Dash, Vineeth~N Balasubramanian, and Amit Sharma.
\newblock Evaluating and mitigating bias in image classifiers: A causal
  perspective using counterfactuals.
\newblock {\em arXiv preprint arXiv:2009.08270}, 2020.

\bibitem{ganin2016domain}
Yaroslav Ganin, Evgeniya Ustinova, Hana Ajakan, Pascal Germain, Hugo
  Larochelle, Fran{\c{c}}ois Laviolette, Mario Marchand, and Victor Lempitsky.
\newblock Domain-adversarial training of neural networks.
\newblock {\em The {J}ournal of {M}achine {L}earning {R}esearch},
  17(1):2096--2030, 2016.

\bibitem{grover2019bias}
Aditya Grover, Jiaming Song, Ashish Kapoor, Kenneth Tran, Alekh Agarwal, Eric~J
  Horvitz, and Stefano Ermon.
\newblock Bias correction of learned generative models using likelihood-free
  importance weighting.
\newblock In {\em Advances in {N}eural {I}nformation {P}rocessing {S}ystems},
  pages 11058--11070, 2019.

\bibitem{hwang2020fairfacegan}
Sunhee Hwang, Sungho Park, Dohyung Kim, Mirae Do, and Hyeran Byun.
\newblock Fairfacegan: Fairness-aware facial image-to-image translation, 2020.

\bibitem{pix2pix2017}
Phillip Isola, Jun-Yan Zhu, Tinghui Zhou, and Alexei~A. Efros.
\newblock Image-to-image translation with conditional adversarial networks.
\newblock In {\em Proceedings of the {IEEE} {C}onference on {C}omputer {V}ision
  and {P}attern {R}ecognition}, pages 1125--1134, July 2017.

\bibitem{karras2019style}
Tero Karras, Samuli Laine, and Timo Aila.
\newblock A style-based generator architecture for generative adversarial
  networks.
\newblock In {\em Proceedings of the {IEEE} {C}onference on {C}omputer {V}ision
  and {P}attern {R}ecognition}, pages 4401--4410, 2019.

\bibitem{kingma2013auto}
Diederik~P Kingma and Max Welling.
\newblock Auto-encoding variational bayes.
\newblock {\em arXiv preprint arXiv:1312.6114}, 2013.

\bibitem{liu2015faceattributes}
Ziwei Liu, Ping Luo, Xiaogang Wang, and Xiaoou Tang.
\newblock Deep learning face attributes in the wild.
\newblock In {\em Proceedings of International Conference on Computer Vision
  (ICCV)}, December 2015.

\bibitem{Locatello2019OnTF}
Francesco Locatello, Gabriele Abbati, Tom Rainforth, S. Bauer, B.
  Sch{\"o}lkopf, and Olivier Bachem.
\newblock On the fairness of disentangled representations.
\newblock In {\em NeurIPS}, 2019.

\bibitem{louizos2015variational}
Christos Louizos, Kevin Swersky, Yujia Li, Max Welling, and Richard Zemel.
\newblock The variational fair autoencoder.
\newblock {\em arXiv preprint arXiv:1511.00830}, 2015.

\bibitem{madras2018learning}
David Madras, Elliot Creager, Toniann Pitassi, and Richard Zemel.
\newblock Learning adversarially fair and transferable representations.
\newblock In {\em International Conference on Machine Learning}, pages
  3384--3393. PMLR, 2018.

\bibitem{paul2020tara}
William Paul, Armin Hadzic, Neil Joshi, Fady Alajaji, and Phil Burlina.
\newblock Tara: Training and representation alteration for ai fairness and
  domain generalization.
\newblock {\em arXiv preprint arXiv:2012.06387}, 2020.

\bibitem{paul2020unsupervised}
William Paul, I-Jeng Wang, Fady Alajaji, and Philippe Burlina.
\newblock Unsupervised semantic attribute discovery and control in generative
  models.
\newblock {\em Neural {C}omputation}, 33(3):802–--826, 2021.

\bibitem{pix2pix2019}
Yanyun Qu, Yizi Chen, Jingying Huang, and Yuan Xie.
\newblock Enhanced pix2pix dehazing network.
\newblock In {\em Proceedings of the {IEEE}/{CVF} {C}onference on {C}omputer
  {V}ision and {P}attern {R}ecognition}, pages 8160--8168, June 2019.

\bibitem{quadrianto2019discovering}
Novi Quadrianto, Viktoriia Sharmanska, and Oliver Thomas.
\newblock Discovering fair representations in the data domain.
\newblock In {\em Proceedings of the {IEEE} {C}onference on {C}omputer {V}ision
  and {P}attern {R}ecognition}, pages 8227--8236, 2019.

\bibitem{sattigeri2018fairness}
Prasanna Sattigeri, Samuel~C. Hoffman, Vijil Chenthamarakshan, and Kush~R.
  Varshney.
\newblock Fairness gan, 2018.

\bibitem{shokri2017membership}
Reza Shokri, Marco Stronati, Congzheng Song, and Vitaly Shmatikov.
\newblock Membership inference attacks against machine learning models.
\newblock In {\em 2017 {IEEE} {S}ymposium on {S}ecurity and {P}rivacy ({SP})},
  pages 3--18. {IEEE}, 2017.

\bibitem{song2019learning}
Jiaming Song, Pratyusha Kalluri, Aditya Grover, Shengjia Zhao, and Stefano
  Ermon.
\newblock Learning controllable fair representations.
\newblock In {\em The 22nd International {C}onference on {A}rtificial
  {I}ntelligence and {S}tatistics}, pages 2164--2173, 2019.

\bibitem{Stallkamp-IJCNN-2011}
Johannes Stallkamp, Marc Schlipsing, Jan Salmen, and Christian Igel.
\newblock The {G}erman {T}raffic {S}ign {R}ecognition {B}enchmark: A
  multi-class classification competition.
\newblock In {\em IEEE International Joint Conference on Neural Networks},
  pages 1453--1460, 2011.

\bibitem{wadsworth2018achieving}
Christina Wadsworth, Francesca Vera, and Chris Piech.
\newblock Achieving fairness through adversarial learning: an application to
  recidivism prediction.
\newblock In {\em {C}onference on {F}airness, {A}ccountability, and
  {T}ransparency in {M}achine {L}earning ({FATML})}, 2018.

\bibitem{xie2020unsupervised}
Qizhe Xie, Zihang Dai, Eduard Hovy, Minh-Thang Luong, and Quoc~V. Le.
\newblock Unsupervised data augmentation for consistency training, 2020.

\bibitem{zhang2018mitigating}
Brian~Hu Zhang, Blake Lemoine, and Margaret Mitchell.
\newblock Mitigating unwanted biases with adversarial learning.
\newblock In {\em Proceedings of the 2018 {AAAI}/{ACM} {C}onference on {AI},
  {E}thics, and {S}ociety}, pages 335--340, 2018.

\bibitem{zhao2017infovae}
Shengjia Zhao, Jiaming Song, and Stefano Ermon.
\newblock Infovae: Information maximizing variational autoencoders.
\newblock {\em arXiv preprint arXiv:1706.02262}, 2017.

\bibitem{cyclegan2017}
Jun-Yan Zhu, Taesung Park, Phillip Isola, and Alexei~A Efros.
\newblock Unpaired image-to-image translation using cycle-consistent
  adversarial networks.
\newblock In {\em Proceedings of the {IEEE} {I}nternational {C}onference on
  {C}omputer {V}ision}, pages 2223--2232, 2017.

\end{thebibliography}
}

\clearpage

\section*{Appendix}

\begin{figure}[t]
\centering 
\includegraphics[page=1,width=\linewidth]{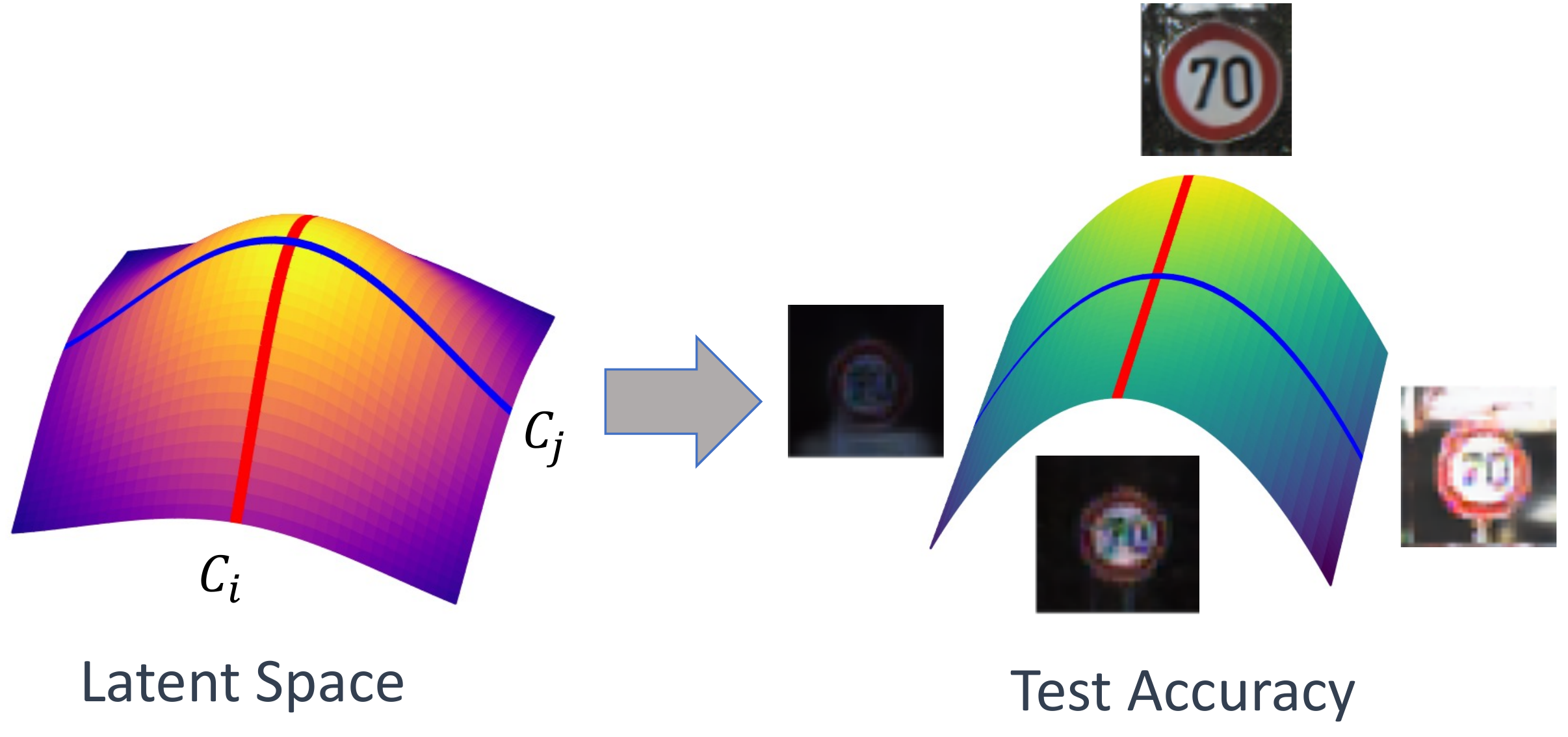}
     \vspace{-0.2in}
    \caption{Main idea: We learn variables such as $C_i$ or $C_j$ in latent space that corresponds to a) factors of variations such as lighting (blue) or sign size (red) and b) to probe or address how sensitive task performance (low performance is purple versus high performance is yellow-green) is to these factors. 
    }
\label{fig:overview1}
\end{figure}

The figure~\ref{fig:overview1} depicts our overall approach in a succinct graphic.

\subsection{Unsupervised Invariance}
Tables \ref{tab:gtsrb_uf} and \ref{tab:celeba_uf} show mor extensive view of our results  from evaluating the sensitivity and intervening with respect to each one of the ten learnt factors of variation to make the end task invariant to these factors. The results in the main text for the unsupervised invariance were taken from these tables corresponding to the column for the stated factor. Note that after the first three factors for GTSRB and the first two factors for CelebA, the accuracy gap drops significantly (from 13.47 to 1.34 for GTSRB and 7.20 to 3.62 for CelebA). These factors were the ones deemed most sensitive and reported in the main text, with the exception of factor 5 for GTSRB which appeared to control similar aspects as factor 7. Thus we only evaluate factor 7 over 5 as it is more sensitive.
\begin{table*}[h]

  \renewcommand{\arraystretch}{1.2}
\scriptsize
    \centering
    \begin{tabular}{c|c|c|c|c|c|c|c|c|c|c|c}
    
	    \toprule
        \multicolumn{2}{c|}{\textbf{Sorted Sensitive Factors}}&1&7&5&9&6&8&10&3&4&2\\
        
          \midrule
        \midrule
        \multirow{2}{*}{\textbf{Baseline}}&Accuracy&98.49&98.49&98.49&98.49&98.49&98.49&98.49&98.49&98.49&98.49\\
&Accuracy Gap&22.93&18.46&13.47&1.34&1.17&1.13&0.95&0.91&0.87&0.71\\
&Min Accuracy&65.80&69.37&72.19&81.93&81.85&81.91&81.81&82.12&81.93&82.09\\
        
        \cline{1-12}
        \multirow{4}{*}{\parbox{1.3cm}{\centering \textbf{Semantic Augmentation}}}&Accuracy&\textbf{98.49}&\textbf{99.06}&\textbf{98.89}&\textbf{98.90}&\textbf{98.65}&\textbf{98.69}&\textbf{98.55}&\textbf{98.86}&\textbf{99.09}&\textbf{99.11}\\
&Accuracy Gap&\textbf{4.12}&\textbf{2.81}&\textbf{2.89}&\textbf{0.52}&\textbf{0.44}&\textbf{0.63}&\textbf{0.47}&\textbf{0.45}&\textbf{0.60}&\textbf{0.23}\\
&Min Accuracy&\textbf{94.29}&\textbf{95.82}&\textbf{95.39}&\textbf{97.49}&\textbf{97.39}&\textbf{95.29}&\textbf{96.87}&\textbf{97.42}&\textbf{97.43}&\textbf{97.67}\\
&CAI(0.5)&\textbf{9.46}&\textbf{8.09}&\textbf{5.49}&\textbf{0.62}&\textbf{0.45}&-\textbf{0.35}&\textbf{0.27}&\textbf{0.42}&\textbf{0.44}&\textbf{0.55}\\
&CAI(0.75)&\textbf{14.14}&\textbf{11.87}&\textbf{8.03}&\textbf{0.72}&\textbf{0.59}&\textbf{0.43}&\textbf{0.37}&\textbf{0.44}&\textbf{0.35}&\textbf{0.51}\\
        \cline{1-12}
        \multirow{4}{*}{\parbox{1.3cm}{\centering \textbf{Adversarial}}}&Accuracy&97.99&98.01&97.30&96.88&97.55&92.93&94.24&97.14&95.68&21.43\\
&Accuracy Gap&10.84&4.45&5.87&0.70&0.80&1.14&0.52&0.65&0.96&1.05\\
&Min Accuracy&86.09&92.00&90.04&90.79&92.87&82.67&86.72&93.93&92.09&12.71\\
&CAI(0.5)&5.77&6.74&3.21&-0.48&-0.29&-2.78&-1.91&-0.54&-1.45&-38.70\\
&CAI(0.75)&8.93&10.38&5.41&0.08&0.04&-1.39&-0.74&-0.14&-0.77&-19.52\\
\cline{1-12}
        \multirow{4}{*}{\parbox{1.3cm}{\centering \textbf{Coherence Regularization}}}&Accuracy&96.31&95.96&95.99&97.35&97.85&97.46&97.43&98.03&98.10&97.92\\
&Accuracy Gap&34.85&15.65&10.86&1.41&1.43&1.11&1.00&0.96&0.78&1.23\\
&Min Accuracy&48.50&65.30&64.18&81.57&81.42&81.22&81.02&80.36&81.27&79.96\\
&CAI(0.5)&-6.97&0.10&0.06&-0.60&-0.45&-0.50&-0.55&-0.25&-0.15&-0.54\\
&CAI(0.75)&-9.44&1.46&1.34&-0.33&-0.35&-0.24&-0.30&-0.15&-0.03&-0.53\\
	    \bottomrule
    \end{tabular}
    \caption{GTSRB Unsupervised Invariance Results. Numbers in bold denote which one of the three interventions performed the best {\bf and} improved over the baseline (for CAI, when it is non-negative). }
    \label{tab:gtsrb_uf}
\end{table*}

\begin{table*}[]

  \renewcommand{\arraystretch}{1.2}
\scriptsize
    \centering
    \begin{tabular}{c|c|c|c|c|c|c|c|c|c|c|c}
    
	    \toprule
        \multicolumn{2}{c|}{\textbf{Sorted Sensitive Factors}}&5&4&6&7&9&2&10&8&1&3\\
        
          \midrule
        \midrule
        \multirow{2}{*}{\textbf{Baseline}}&Accuracy&81.34&81.34&81.34&81.34&81.34&81.34&81.34&81.34&81.34&81.34\\
&Accuracy Gap&8.77&7.20&3.62&3.59&3.38&2.15&1.87&1.67&1.57&1.06\\
&Min Accuracy&69.80&70.09&72.00&71.60&72.28&72.76&72.81&73.03&73.03&73.20\\
        
        \cline{1-12}
        \multirow{4}{*}{\parbox{1.3cm}{\centering \textbf{Semantic Augmentation}}}&Accuracy&\textbf{79.45}&\textbf{82.53}&79.67&79.17&79.80&\textbf{78.89}&78.93&78.53&78.93&79.84\\
&Accuracy Gap&\textbf{1.02}&\textbf{1.05}&\textbf{1.44}&1.07&1.38&\textbf{0.61}&\textbf{1.39}&\textbf{0.96}&0.90&1.70\\
&Min Accuracy&\textbf{86.34}&\textbf{86.19}&\textbf{86.40}&\textbf{86.03}&\textbf{85.69}&\textbf{86.51}&\textbf{85.94}&\textbf{86.24}&\textbf{86.22}&\textbf{85.96}\\
&CAI(0.5)&\textbf{2.89}&1.85&\textbf{0.31}&0.23&0.28&-0.40&-0.91&-1.00&-0.82&-1.02\\
&CAI(0.75)&\textbf{5.32}&\textbf{4.00}&\textbf{1.25}&1.38&1.14&\textbf{0.57}&-0.22&-0.15&-0.07&-0.83\\
        \cline{1-12}
        \multirow{4}{*}{\parbox{1.3cm}{\centering \textbf{Adversarial}}}&Accuracy&74.11&80.12&67.65&50.03&70.42&65.46&63.71&76.16&62.48&59.89\\
&Accuracy Gap&1.36&3.99&4.70&1.56&2.66&1.55&5.17&2.46&3.49&5.92\\
&Min Accuracy&84.37&81.59&56.78&57.66&79.62&74.50&61.56&71.12&70.16&68.19\\
&CAI(0.5)&0.14&0.99&-7.33&-14.59&-5.05&-7.59&-10.41&-2.94&-10.34&-13.10\\
&CAI(0.75)&3.78&2.10&-4.20&-6.28&-2.16&-3.50&-6.85&-1.87&-6.13&-8.98\\
\cline{1-12}
\multirow{4}{*}{\parbox{1.3cm}{\centering \textbf{Coherence Regularization}}}&Accuracy&80.38&81.48&\textbf{80.79}&\textbf{81.38}&\textbf{79.79}&78.24&\textbf{81.26}&\textbf{81.83}&81.11&\textbf{80.81}\\
&Accuracy Gap&7.86&2.12&2.66&\textbf{0.91}&\textbf{0.97}&2.85&1.63&1.75&0.96&3.10\\
&Min Accuracy&67.94&69.48&68.58&71.50&65.50&64.59&70.20&71.09&73.52&70.77\\
&CAI(0.5)&-0.02&\textbf{2.58}&0.26&\textbf{1.41}&\textbf{0.48}&-1.85&\textbf{0.13}&\textbf{0.25}&\textbf{0.25}&-1.23\\
&CAI(0.75)&0.45&3.83&0.61&\textbf{2.05}&\textbf{1.44}&-1.28&\textbf{0.18}&\textbf{0.08}&\textbf{0.43}&-1.64\\
	    \bottomrule
    \end{tabular}
    \caption{CelebA Unsupervised Invariance Results. Numbers in bold denote which one of the three interventions performed the best {\bf and} improved over the baseline (for CAI, when it is non-negative). }
    
    \label{tab:celeba_uf}
\end{table*}
\end{document}